\documentclass[letterpaper]{article} 
\usepackage{aaai25}
\usepackage{array}
\usepackage{multirow}
\usepackage{float}
\usepackage{makecell}
\usepackage{times}  
\usepackage{helvet, amsmath}  
\usepackage{courier,enumerate}  
\usepackage[hyphens]{url}  
\usepackage{graphicx} 
\usepackage{subcaption}
\usepackage{natbib}  
\usepackage{caption} 

\newcommand{\bm}{\boldsymbol}
\newcommand{\Xpar}{\mathbb X_{\sf pareto}}

\newcommand{\pubPG}{PUB-MOBO-PG}
\newcommand{\pubPGoe}{PUB-MOBO-PG+OE}

\newcommand{\eubo}{EUBO+qEIUU}
\newcommand{\xgd}{\bm x_{\text{GD}}}
\newcommand{\xexp}{\bm x_{\text{EXP}}}

\usepackage{algorithm}
\usepackage{algorithmic}

\usepackage{amsmath}
\usepackage{amssymb}
\usepackage{xcolor}

\usepackage{newfloat}
\usepackage{listings}
\DeclareCaptionStyle{ruled}{labelfont=normalfont,labelsep=colon,strut=off}
\floatstyle{ruled}
\newfloat{listing}{tb}{lst}{}
\floatname{listing}{Listing}

\newtheorem{assumption}{Assumption}

\title{User Preference Meets Pareto-Optimality in Multi-Objective Bayesian Optimization}

\author{
    Joshua Hang Sai Ip\textsuperscript{\rm 1},
    Ankush Chakrabarty\textsuperscript{\rm 2},
    Ali Mesbah\textsuperscript{\rm 1},
    Diego Romeres\textsuperscript{\rm 2}
}
\affiliations{
    \textsuperscript{\rm 1}University of California, Berkeley\\
    \textsuperscript{\rm 2}Mitsubishi Electric Research Laboratories\\
    ipjoshua@berkeley.edu, chakrabarty@merl.com, mesbah@berkeley.edu, romeres@merl.com
}

\DeclareMathOperator*{\argmax}{arg\,max}

\begin{document}

\maketitle

\begin{abstract}
Incorporating user preferences into multi-objective Bayesian optimization (MOBO) allows for personalization of the optimization procedure.
Preferences are often abstracted in the form of an unknown utility function, estimated through pairwise comparisons of potential outcomes.
However, utility-driven MOBO methods can yield solutions that are dominated by nearby solutions, as non-dominance is not enforced.
Additionally, classical MOBO commonly relies on estimating the entire Pareto-front to identify the Pareto-optimal solutions, which can be expensive and ignore user preferences.
Here, we present a new method, termed preference-utility-balanced MOBO (PUB-MOBO), that allows users to disambiguate between near-Pareto candidate solutions.
PUB-MOBO combines utility-based MOBO with local multi-gradient descent to refine user-preferred solutions to be near-Pareto-optimal.
To this end, we propose a novel preference-dominated utility function that concurrently preserves user-preferences and dominance amongst candidate solutions.
A key advantage of PUB-MOBO is that the local search is restricted to a (small) region of the Pareto-front directed by user preferences, alleviating the need to estimate the entire Pareto-front.
PUB-MOBO is tested on three synthetic benchmark problems: DTLZ1, DTLZ2 and DH1, as well as on three real-world problems: Vehicle Safety, Conceptual Marine Design, and Car Side Impact. PUB-MOBO consistently outperforms state-of-the-art competitors in terms of proximity to the Pareto-front and utility regret across all the problems. 
\end{abstract}

\section{Introduction}
The challenge of identifying optimal trade-offs between multiple complex objective functions is pervasive in many real-world scientific and industrial applications. When the objectives are black-box functions constructed from noisy observations, multi-objective Bayesian optimization (MOBO) \cite{konakovic2020diversity,daulton2020differentiable} is an effective multi-objective optimization (MOO) approach owing to its high sample efficiency, especially as compared to classic MOO methods such as CMA-ES \cite{hansen2001completely} and NSGA-II \cite{deb2002fast}. Nonetheless, state-of-the-art MOBO methods generally seek to estimate the entire Pareto-front, which can become prohibitively expensive due to sample-based evaluation of acquisition functions such as $q$-Expected Hypervolume Improvement (qEHVI) \cite{daulton2020differentiable}.    

Recently, there has been a growing interest in preference-based MOBO (e.g., \cite{abdolshah2019multi,ahmadianshalchi2024preference, shao2023preference, ozaki2024multi}) that aims to incorporate user preferences into MOO for selecting optimal points. In essence, preference-based MOBO guides the optimization process towards regions of interest within the Pareto-front by leveraging user feedback, typically in the form of pairwise comparisons between solutions generated by the optimization algorithm. These comparisons are used to estimate an underlying utility function that describes user preferences. However, while preference-based MOBO can effectively identify solutions with high utility as informed by user feedback, the resulting solutions may not be Pareto-optimal.

To address this challenge, we present a new method, termed Preference-Utility-Balanced MOBO (PUB-MOBO), that systematically approaches the user-informed regions of interest within the Pareto-front by synergizing global and local search strategies. PUB-MOBO begins with a global search driven by utility maximization to identify regions in the solution space that align with user preferences. Subsequently, a local search is conducted in the vicinity of these solutions to discover solutions that are closer to Pareto-optimality. Additionally, a new utility function, the Preference-Dominated Utility Function (PDUF), is proposed that encapsulates the concept of dominance within a single function. PDUF allows for consistently identifying dominating solutions, while providing a straightforward means for expressing all possible user preferences. This differs to existing utility functions for preference-based MOBO such as the negative  $l_1$  distance from an ideal solution irrespective of the solution being on the Pareto-front or an infeasible ideal solution \cite{miettinen1999nonlinear}, or the weighted sum where not all Pareto-optimal points can be assigned with the highest utility value from any choice of weights \cite{chiandussi2012comparison}. Empirical demonstrations on several synthetic benchmark and real-world problems  show that PUB-MOBO not only enhances the utility of the optimization solutions, but also yields near Pareto-optimal solutions.  

In sum, the main contributions of this paper include:
\begin{enumerate}[(C1)]
\item We introduce a new preference-based MOBO method, PUB-MOBO, that can effectively integrate feedback about user preferences and moves towards Pareto-optimality in MOO of black-box functions.
\item We propose the use of gradient descent (GD) in the context of MOBO, along with a new utility function, PDUF, that seamlessly combine user preferences with the notion of dominance to identify user-preferred solutions that are approximately Pareto-optimal. 
\item We illustrate the importance of reducing gradient uncertainty in GD to aid PUB-MOBO in locating high-utility near Pareto-optimal solutions. Our approach is based on the Gradient Information acquisition function \cite{muller2021local}.

\item Through numerical experiments on synthetic and real-world benchmark problems, we demonstrate that incorporating GD into utility-based MOBO can significantly enhance performance, yielding solutions that better align user preferences and Pareto-optimality.
\end{enumerate}

\section{Related Work}
Traditional MOBO methods such as $q$-EHVI \cite{daulton2020differentiable} assume that all Pareto-optimal solutions are equally desirable to the user, which might not be the case in practice. Instead, preference-based MOBO models user preferences with utility functions. \citet{lin2022preference,astudillo20a} propose the EUBO and qEIUU acquisition functions respectively, which take advantage of user-preference when querying new points.

Various works have proposed different methods to incorporate gradients as additional information in single-objective BO to enhance global search. \citet{wu2017gradient} construct a joint GP to correlate zeroth and first order information, demonstrating that gradients aid the surrogate in approximating the posterior and change which points to query. Similarly, \citet{makrygiorgos2023no} leverage gradients to establish stationarity conditions within a Karush-Kuhn-Tucker (KKT) formulation.

Other works instead use gradients for first-order optimization, also in the single-objective BO context. Bayesian and local optimisation sample-wise switching optimisation method (BLOSSOM) switches from global search to local search with BFGS when the posterior objective is close to the objective \cite{mcleod2018optimization}. On the other hand, \cite{muller2021local, nguyen2022local} abandon global search and construct local GPs for local optimization. 
To our knowledge, involving gradients for preference-based MOBO has not been studied and PUB-MOBO is the first algorithm to do this.

\section{Preliminaries}
\subsection{Problem Formulation}

We tackle a multi-objective optimization (MOO) problem for \textit{minimizing} $n_f$ expensive-to-evaluate objective functions, denoted by $f_i(\bm x)$ for $i\in \{1,\cdots, n_f\}$. Consequently, the objective function vector is denoted $\bm f(\bm x)$, where $\bm x\in\mathbb{R}^{n_x}$ denote the decision variables. We assume that for a candidate $\bm x$, the function $\bm f(\bm x)$ can be evaluated, but no first- or higher-order information about any component of $\bm f$ is available. Furthermore, an analytical form of $\bm f$ is not known. 

For MOO problems without user-preferences, the objective is to attain Pareto-optimality, which is defined as follows~\cite{deb2005searching}. A point $\mathbf{\bar{x}}$ in a feasible set $\mathbb{X}$ is \textit{Pareto-optimal} if it is not dominated\footnote{Generally, a candidate $\bm x_1$ \textit{dominates} a candidate $\bm x_2$, denoted by $\bm x_1 \succ \bm x_2$, if 
 $f_i(\bm x_1) \le f_i(\bm x_2)$ for all $i=1,\ldots, n_f$ and for one such element, $f_i(\bm x_1) < f_i(\bm x_2)$.} by any other solution in $\mathbb X$. That is, there exists no other feasible point $\mathbf{x} \in \mathbb{X}$ such that $f_i(\bm {x}) \leq f_i(\bm{\bar{x}})$ for all $i \in \{1,\cdots, n_f\}$ and $f_j(\bm{x}) < f_j(\bm {x}^\star)$ for at least one $j \in \{1,\cdots, n_f\}$. Note that accurately computing the set of Pareto-optimal points, referred to as the Pareto-front $\Xpar$, can often be computationally prohibitive, even for small $n_f$. 

In the presence of a user,  estimating the entire Pareto-front may become unnecessary, especially when only specific sub-regions of the feasible set $\mathbb{X}$ is of interest. Mathematically, such user-preferences are often abstracted in the MOBO literature via \textit{utility functions}. Specifically, the MOO problem is recast as a (scalar) utility maximization problem
\begin{equation}\label{eq:PGBO}
    \max_{\bm{x}\in \mathbb{X}} \quad u\left(\bm {f}(\bm {x})\right),
\end{equation}
where $u: \mathbb{R}^{n_f} \to \mathbb{R}$ is the unknown utility function that dictates the behavior of the user. Note that the input to the utility is a noise-corrupted outcome vector $\bm {y} = \bm {f}(\bm {x})+\varepsilon$, where $\varepsilon$ is zero-mean noise  
with variance $\sigma^2_\varepsilon \bm I_{n_y}$ where $\bm I_{n_f}$ is the $n_f\times n_f$ identity matrix. 
For the sequel, let the highest utility Pareto-point be defined as
\begin{equation}\label{eq:x-best}
\bm x^* \in \argmax_{\bm x\in \Xpar} u(\bm f(\bm x)).
\end{equation}
Following the preference-based BO literature, we assume that the utility function is not available to evaluate directly, and no functional form is known. Additionally, it is well-established that user preferences are difficult to be assigned to continuous numerical values; instead we suppose that users are more inclined to provide weak supervision in the form of pairwise comparisons~\cite{chu2005preference, lin2022preference}.  
The following assumption asserts that a typical user will select dominating solutions when possible. 
\begin{assumption}\label{asmp:1}
If $\bm y_1$ and $\bm y_2$ are candidate outcomes presented to the user and $\bm y_1\succ\bm y_2$, then the user will always select $\bm y_1$; that is, $u(\bm y_1) > u(\bm y_2)$.
\end{assumption}
This assumption is a constraint that must be respected when modeling preference-based MOBO problems to accurately reflect real user behavior.

\subsection{Modeling with Gaussian Processes}
We first discuss the modeling choices considered to learn the outcome function $\boldsymbol{f}$ and the utility function $u$: their respective approximations are denoted $\hat{\bm f}$ and $\hat u$. 
\subsubsection{Modeling outcomes.}
Gaussian process (GP) regression is a popular choice for constructing the surrogate $\boldsymbol{\hat{f}}$ for the true outcome function $\boldsymbol{f}$. We train an independent GP for each objective $\boldsymbol{f}_i$, though a multi-output GP that models correlations between the objectives could also be considered \cite{alvarez2012kernels}. Each GP is defined \textit{a priori} by a mean function $m(\boldsymbol{x})$ and covariance function $k_i(x, x')$ called kernel. For this work, any $\mathcal C^2$ kernel is admissible.

Let $\bm X_T = [\bm x_1, \bm x_2, \ldots, \bm x_T]$; we drop the subscript for brevity. Given a dataset $D :=(\boldsymbol{X}, \boldsymbol{Y})$, comprising input-outcome pairs, the mean and variance of the posterior are:
\begin{subequations}
\label{eq: GP_inference}
\begin{align}
\label{eq: GP_inference_mean}
\mu_i(\boldsymbol{x}) &= m(\boldsymbol{x})
+ k_i(\boldsymbol{x}, \boldsymbol{X})\mathcal K_\sigma^{-1}(\boldsymbol{X})(Y_i-m(\boldsymbol{X})),\\
\label{eq: GP_inference_var}
\Sigma_i(\boldsymbol{x}) &= k_i(\boldsymbol{x}, \boldsymbol{x}) 
- k_i(\boldsymbol{x}, \boldsymbol{X})\mathcal K_\sigma^{-1}(\boldsymbol{X})k_i(\boldsymbol{X}, \boldsymbol{x}),
\end{align}
\end{subequations}
where $\mathcal K_\sigma(\bm X):= K(\bm X, \bm X) + \sigma^2\,\bm I$ and $m(\cdot)$ is the prior mean.
Since the derivative is a linear operator, the derivative GP is another GP~ \cite{williams2006gaussian} characterized fully by the mean and covariance functions
\begin{subequations}
\label{eq: dGP_inference}
\begin{align}
\label{eq: dGP_inference_mean} 
\mu^\nabla_i(\boldsymbol{x}) &= \nabla m(\boldsymbol{x}) 
+ \nabla k_i(\boldsymbol{x}, \boldsymbol{X})\mathcal K_\sigma^{-1}(\boldsymbol{X})(Y_i-m(\boldsymbol{X})),\\
\label{eq: dGP_inference_covar}\Sigma^\nabla_i(\boldsymbol{x}) &= \nabla^2k_i(\boldsymbol{x}, \boldsymbol{x}) 
- \nabla k_i(\boldsymbol{x}, \boldsymbol{X})\mathcal K_\sigma^{-1}(\bm{X})\nabla k_i(\boldsymbol{X}, \boldsymbol{x}),
\end{align}
\end{subequations}

\subsubsection{Modeling Preferences.} \label{subsubsec:modeling_preferences}
We assume the user is only capable of weak supervisions in the form of pairwise comparisons (PC). That is, if the user prefers $\bm y:=\boldsymbol{f}$ over $\bm y:=\boldsymbol{f'}$, the pairwise comparison function $r(\bm y, \bm y')=0$. In the event that the user prefers $\bm y'$ instead,  $r(\bm y, \boldsymbol{y'})=1$. Pairwise GPs c.f.~\cite{chu2005preference} allow us to learn a latent functional representation $\hat u$ of the true user utility based on this preference feedback. The latent function satisfies $\hat{u}(\boldsymbol{y})>\hat{u}(\boldsymbol{y'})$ if the user prefers $\boldsymbol{y}$, and vice versa.

\section{Preference-Utility-Balanced (PUB) MOBO}

This work is motivated by the practical consideration that most users expect to be shown promising candidate outcomes after very few interactions. Furthermore, they often require some assurance that the suggested candidates are not only high in utility, but also high in performance, i.e., close to the Pareto-front. To this end, we propose PUB-MOBO, which relies on utility maximization to ascertain candidate solutions that are preferred by the user, while promoting a local search towards the Pareto-front using estimated gradients. Empirically, we observe that the local search finds solutions near Pareto points, which subsequently accelerates the search for high-utility solutions. The rationale behind this is that the local search likely yields dominating solutions over those found by utility maximization alone.
These dominating solutions, by Assumption~\ref{asmp:1}, have higher utility. This implies that even with few user interactions, i.e. pairwise comparisons, we can obtain promising candidates that are unlikely to be dominated. As more user feedback is collected, we expect to obtain near-maximal utility solutions due to the utility maximization, from which following local gradients should result in a near-Pareto solution. 

\subsection{PUB-MOBO Algorithm}
The proposed PUB-MOBO method operates in three stages. Following the work of~\citet{lin2022preference}, the first two stages of the framework are the preference exploration (PE) stage and the outcome evaluation via experimentation (EXP) stage. Our method extends these two stages with an additional stage based on local multi-gradient descent, denominated the \textit{GD stage}. 

In each PUB-MOBO iteration, these three stages are executed, and the process is repeated \textit{ad infinitum}, or (more practically) until a pre-decided budget for total number of outcome/function evaluations is attained; see Algorithm \ref{alg:PUB-MOBO}.

\begin{algorithm}[tb]
\caption{\textsc{PUB-MOBO}}
\label{alg:PUB-MOBO}
\begin{algorithmic}[1] 
\STATE Generate initial data: $\bm{x_{\text{INIT}}}, \bm{y_{\text{INIT}}}, r(\bm{y_{\text{INIT}}})$
\STATE $D = (\bm x_{\text{INIT}}, \bm y_{\text{INIT}})$
\STATE $P\leftarrow $ comparisons on a subset of $\bm{y_{\text{INIT}}} \times \bm{y_{\text{INIT}}}$
\STATE Update outcome model $\boldsymbol{\hat{f}} \ \text{with}\ D$
\STATE Update preference model $\hat{u}\ \text{with}\ P$
\WHILE{\# outcome evaluations $\le$ budget}
\STATE \textbf{PE stage}
\STATE $\boldsymbol{x_{1}}, \boldsymbol{x_{2}} \leftarrow$ $\text{argmax}_{\boldsymbol{x_1}, \boldsymbol{x_2}}$ EUBO
\STATE $\boldsymbol{y_{1}},\ \boldsymbol{y_{2}} = \boldsymbol{\hat{f}(x_{1})}, \boldsymbol{\hat{f}(x_{2})}$
\STATE $r(\boldsymbol{y_{1},\ y_{2}}) \leftarrow$ user provides a comparison
\STATE Append $P$ with $(\boldsymbol{y_{1},\ y_{2}}, r(\boldsymbol{y_{1},\ y_{2}}))$
\STATE Update pref. model $\hat{u}$ with $(\boldsymbol{y_{1},\ y_{2}}, r(\boldsymbol{y_{1},\ y_{2}}))$
\STATE \textbf{EXP stage}
\STATE $\boldsymbol{x_{\text{EXP}}} \leftarrow$ $\text{argmax}_{\boldsymbol{x}}$ qEIUU
\STATE $\boldsymbol{y_{\text{EXP}}} = \boldsymbol{f(x_{\text{EXP}})}$
\STATE Append $D$ with $(\boldsymbol{x_{\text{EXP}}}, \boldsymbol{y_{\text{EXP}}})$
\STATE Update outcome model $\boldsymbol{\hat{f}}$ with $(\boldsymbol{x_{\text{EXP}}}, \boldsymbol{y_{\text{EXP}}})$
\STATE \textbf{GD stage}
\STATE $(\bm X_{\text{GD}}, \bm Y_{\text{GD}}) \leftarrow$ Local Gradient Descent($\boldsymbol{x_{\text{EXP}}}$) 
\STATE Append $D$ with $(\bm X_{\text{GD}}, \bm Y_{\text{GD}})$
\ENDWHILE
\end{algorithmic}
\end{algorithm}

\subsubsection{Preference Exploration.}
In the PE stage, the user expresses their preferences over a query of two candidate solutions in a form of pairwise comparisons. The comparison is then used to update the estimate of the utility function $\hat{u}$. The candidate solutions in the query are obtained by optimizing the expected utility of both outcomes acquisition function, which is a look-ahead function that maximizes the difference in expected utility after a user query $r(\boldsymbol{\hat{f}(x_1)}, \boldsymbol{\hat{f}(x_2)})$. Since this is expensive to compute, a simplified acquisition called EUBO, proposed in \cite{lin2022preference}, is used instead:
\begin{equation}
\label{eq: EUBO}
\mathsf{EUBO}(\boldsymbol{x_1}, \boldsymbol{x_2}) =\mathbb{E}[\max(\hat{u}(\boldsymbol{\hat{f}}(\boldsymbol{x_1}),\hat{u}(\boldsymbol{\hat{f}}(\boldsymbol{x_2}))].
\end{equation}
This has the advantage of estimating utility with outcome posteriors that are feasible under $\boldsymbol{f}$. Note that no evaluation of $\bm f$ is required in this stage. 

\subsubsection{Outcome Evaluation via Experiments.}
The EXP stage involves computing the optimal decision variables and collecting outcome evaluations to update the outcome model \(\boldsymbol{\hat{f}}\). In this process, the well-established expected improvement under utility uncertainty (qEIUU) acquisition function, \cite{astudillo20a}, widely used in preference-based contexts, is maximized to determine the optimal decision variables, $\bm x_{\text{EXP}}$. Here,
\begin{equation}
\label{eq: EIUU}
\mathsf{qEIUU}(\boldsymbol{x}) = \mathbb{E}\left[\max(\hat{u}(\boldsymbol{\hat{f}(x)})-\hat{u}(\boldsymbol{f(x_{\text{best}})}), 0)\right],
\end{equation}
where $\boldsymbol{x_{\text{best}}} = \arg\max_{x \in \bm X } \hat{u}(\boldsymbol{\hat{f}(x)})$, and $$\xexp:=\argmax_{\mathbb X} \mathsf{qEIUU}(\boldsymbol{x}).$$ Since the expectation in~\eqref{eq: EIUU} is with respect to the outcome and utility models, the analytical expression is challenging to compute. Instead, this can be evaluated via the Monte Carlo approach, where the reparameterization trick is applied to both $\boldsymbol{\hat{f}}$ and $\hat{u}$ and the acquisition function is optimized using sample-averaging; c.f. \citet{wilson2018maximizing, balandat2020botorch}. Note that after $\xexp$ is obtained, we append it along with its true outcome value $\bm f(\xexp)$ to the current dataset $D$.
\subsubsection{Local Gradient Descent.}

This GD stage is motivated by the fact that $\xexp$, while expected to be high in utility, is not specifically designed to be near the Pareto-front. Analogous to single-objective optimization, we will pursue local gradients that are expected to generate a trajectory of $\bm x$ candidates that evolves towards a nearby Pareto-optimal point. We will refer to these gradient-following decision variables as `$\xgd$'. We set the initial $\xgd$ to be $\xexp$.

Clearly, for a MOO problem, gradient descent must be adapted for multiple objectives. We propose the use of multiple gradient descent algorithm (MGDA) \cite{desideri2012multiple}, which was designed for smooth multi-outcome objective functions. While MGDA is provably convergent for white-box optimization problem settings i.e. when gradients of each outcome of $\bm f$ is accessible, it has not been tested in the context of black-box MOO such as in this paper, where gradients are not accessible, and must be estimated. However, MGDA exhibits some theoretical properties that, we hypothesize, and demonstrate via experiments, are beneficial in the MOBO context. 
MGDA exploits the KKT conditions \cite{fliege2000steepest, schaffler2002stochastic}
\begin{equation}
\label{eq: MGDA_KKT}
\boldsymbol{\alpha}\geq \textbf{0}, \;
\textbf{1}^\top\boldsymbol{\alpha} = 1, \;
\boldsymbol{\alpha}^\top\nabla\boldsymbol{f(x)}=0,
\end{equation}
and recasts this for MOO as a quadratic cost constrained on the probability simplex, that is:
\begin{align}
\label{eq:MGDA}
\min_{\boldsymbol{\alpha\ge 0}}& \ \left\|\bm{\alpha}^\top\nabla\bm{f(x)}\right\|^2
\text{ subject to: } \textbf{1}^\top\boldsymbol{\alpha} = 1.
\end{align}
It is well-known, c.f.~\citet{desideri2012multiple}, that a solution to \eqref{eq:MGDA} is either: $\bm{\alpha}^\top\bm{\nabla f(x)}=0$, in which case the current parameters $\boldsymbol{x}$ are Pareto-optimal, or $\bm{\alpha}^\top\bm{\nabla f(x)}\neq 0$, and $\boldsymbol{\alpha}^\top \nabla\boldsymbol{f(x)}$ is a feasible descent direction.
Given that~\eqref{eq:MGDA} is a quadratic cost over linear constraints, we can use the Frank-Wolfe algorithm \cite{sener2018multi,jaggi2013revisiting} to efficiently compute optimal solutions. 

\begin{algorithm}[tb]
\caption{\textsc{Local Gradient Descent}}
\begin{algorithmic}[1]
\label{alg: Local}
\STATE Initialize $\boldsymbol{x_{\text{GD}}} \leftarrow \boldsymbol{x_{\text{EXP}}}$
\STATE $(\bm X_{\text{GD}}, \bm Y_{\text{GD}}) = (\emptyset, \emptyset)$ 
\STATE \# of multi-gradient steps, $n_{\text{GD}}$
\STATE \# of GI optimizations, $n_{\text{GI}}$
\STATE Early stopping threshold, $\varepsilon_{\text{GD}}$
\FOR{$i \leq n_{\text{GD}} 
$}
    \STATE Compute $\bm \mu^\nabla(\xgd)$ using~\eqref{eq: dGP_inference_mean}
        \STATE Compute $\bm M = \bm \mu^\nabla(\xgd)^\top \bm \mu^\nabla(\xgd)$ 
    \STATE $\boldsymbol{\alpha} \leftarrow \textrm{Frank-Wolfe}(\bm{M})$
    \STATE $\boldsymbol{x_{\text{GD}}} \leftarrow \boldsymbol{x_{\text{GD}}} - \eta\boldsymbol{\alpha}^\top\bm \mu^\nabla(\xgd)$ \label{Step:alg_local_pred_grad}
    \IF{$\xgd \in \mathbb X$ \textbf{and} $\left|\left|\boldsymbol{\alpha}^\top\bm \mu^\nabla(\xgd)\right|\right|_2^2 > \varepsilon_{\text{GD}}$}
        \STATE Evaluate the true objective: $\boldsymbol{y_{\text{GD}}} = \boldsymbol{f(x_{\text{GD}})}$
        \STATE Append $(\bm X_{\text{GD}}, \bm Y_{\text{GD}}) \ \text{with}\ (\bm x_{\text{GD}}, \bm y_{\text{GD}})$
        \STATE Update outcome model $\boldsymbol{\hat{f}}\ \text{with}\ \boldsymbol{(x_{\text{GD}}},\boldsymbol{ y_{\text{GD}})}$
        \FOR{$j \leq n_{\text{GI}} 
        $}
        \STATE $\boldsymbol{x_{\text{GI}}} \leftarrow \argmax_{\boldsymbol{x'}}$GI

        \STATE Evaluate the true objective: $\boldsymbol{y_{\text{GI}}} = \boldsymbol{f(x_{\text{GI}})}$
        \STATE Append $(\bm X_{\text{GD}}, \bm Y_{\text{GD}}) \ \text{with}\ (\bm x_{\text{GI}}, \bm y_{\text{GI}})$
        \STATE Update outcome model $\boldsymbol{\hat{f}}\ \text{with}\ \boldsymbol{(x_{\text{GI}}},\boldsymbol{ y_{\text{GI}})}$
        \ENDFOR
    \ELSE
        \STATE \textbf{break}
    \ENDIF
\ENDFOR
\STATE \textbf{return} $(\bm X_{\text{GD}}, \bm Y_{\text{GD}})$
\end{algorithmic}
\end{algorithm}

Solving~\eqref{eq:MGDA} yields an optimal $\bm \alpha$ with which we can take a gradient step $
\xgd \leftarrow \xgd -\eta\boldsymbol{\alpha}^\top\bm \nabla f(\xgd)$.
However, there are two clear difficulties at this juncture. First, this update may yield an $\xgd\not\in \mathbb X$. To counter this, we stop updating when this happens, and stop the local gradient search phase, moving on to the next PUB-MOBO iterations with an updated dataset $D$ that contains all the $\xgd$ and correspond $\bm y_{\text{GD}}$ observed so far. 

The second and more debilitating problem is that we do not have access to gradients of $\bm f$. Thankfully, we do have a surrogate model $\boldsymbol{\hat{f}}$ with which we can obtain an estimate of the gradient at any $\bm x$ with $\bm\mu^\nabla:=\mathbb{E}[\boldsymbol{\nabla\hat{f}(x)}]$ through~\eqref{eq: dGP_inference_mean}. The  gradient step is then
\begin{equation}
\label{eq:grad_step_mgda}
\xgd \leftarrow \xgd -\eta\boldsymbol{\alpha}^\top\bm \mu^\nabla(\xgd).
\end{equation}
Unfortunately, there is no clear correlation between the uncertainties in $\bm f$ and $\boldsymbol{\nabla f}$, so $\boldsymbol{\mu}^\nabla$ could  have large uncertainties even near previously observed points. Therefore, it is imperative to incorporate techniques that can reduce uncertainty in the posterior of the gradient estimate. To this end, we propose to use the gradient information (GI) acquisition function, described in~\citet{muller2021local}.

We explain briefly the mechanism of the GI acquisition. Suppose we select the best candidate from the EXP stage, $\bm x_{\text{EXP}}$, and set it as the initial candidate for the local gradient search: $\bm x_{\text{GD}}$. The GI acquisition tries to select a subsequent point $\bm x'$ that will minimize the uncertainty of the gradient at $\xgd$ if $\bm x'$ and its corresponding $\bm y'$ were known.

By considering all $n_f$ objective independently distributed, we can formulate the uncertainty information contained in the covariance matrix $\bm \Sigma^\nabla$ in~\eqref{eq: dGP_inference_covar} using A-optimal design \cite{chakrabarty2013model}, and maximize:

\begin{equation}
\label{eq: GI_original}
\sum_{i=1}^{n_f} \mathbb{E}\left[\mathsf{Tr}(\Sigma^\nabla_i(\boldsymbol{x_{\text{GD}}}|D))-\mathsf{Tr}\left(\Sigma^\nabla_i\left(\boldsymbol{x_{\text{GD}}}|D,(\boldsymbol{x'}, \boldsymbol{y'})\right)\right)\right],
\end{equation}
which, for Gaussian distributions,  is equivalent to 
\begin{equation}
\label{eq: GI}
\text{GI}(\bm x') = \sum_{i = 1}^{n_f}\textsf{Tr}\left(\nabla k_i(\boldsymbol{x_{\text{GD}}}, \boldsymbol{X'})\mathcal K_\sigma^{-1}(\boldsymbol{X'})\nabla k_i^\top(\boldsymbol{x_{\text{GD}}}, \boldsymbol{X'})\right),
\end{equation}
where $\boldsymbol{X'}=\{\boldsymbol{X} \cup \boldsymbol{x'}\}$. For each gradient-step in $n_\text{GD}$, the GI acquisition function is optimized $n_{\text{GI}}$ times to reduce gradient uncertainty. Upon each optimization, we evaluate the outcome function to obtain a corresponding $\bm y_{\text{GD}}$, which is appended to the dataset $D$ for subsequent PUB-MOBO iterations.

\section{Preference-Dominated Utility Function}
The utility function represents the user preference in preference-based MOBO algorithms and is used to simulate user responses. Practically, the utility function is employed to respond to user queries, such as providing pairwise comparisons between two outcomes \citep{lin2022preference}. A utility function used to test preference-based MOBO algorithms should satisfy two key properties:
\begin{enumerate}[(P1)]
\item \textit{Dominance Preservation}: When evaluating a query, the true utility function should satisfy Assumption~1.
\item \textit{Preference Integration}:
The utility function should have parameters $\theta_u$ that allow unique strictly maximal-utility Pareto-optimal solutions. That is, for any $\bm x\in\Xpar$, there exists an easily computable $\theta_u\in\mathbb{R}^{n_u}$ such that $u(\bm f(\bm x)|\theta_u) > u(\bm f(\{\Xpar\setminus\bm x\}|\theta_u)$.
\end{enumerate}

\begin{figure}[ht]
    \centering
    \begin{subfigure}[b]{0.505\columnwidth}
        \centering
        \includegraphics[width=1.\textwidth]{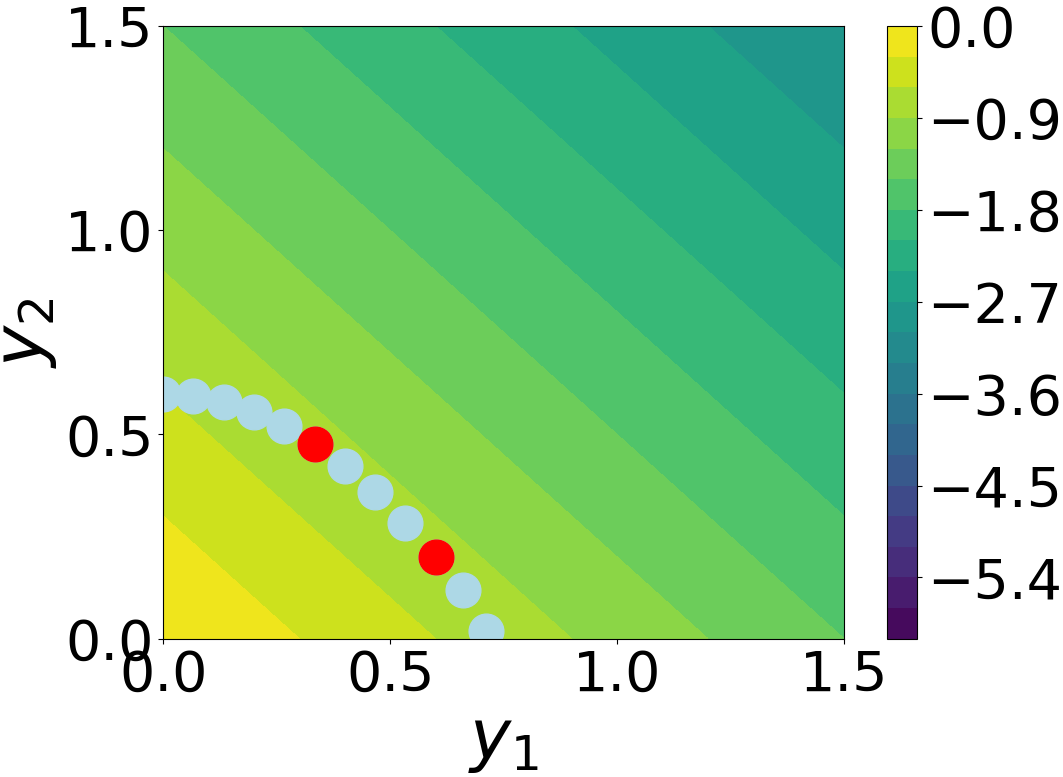}
        \caption{$\ell_1$ distance Utility function}
        \label{fig:l1_utility}
    \end{subfigure}
    \begin{subfigure}[b]{0.48\columnwidth}
        \centering
        \includegraphics[width=1.\textwidth]{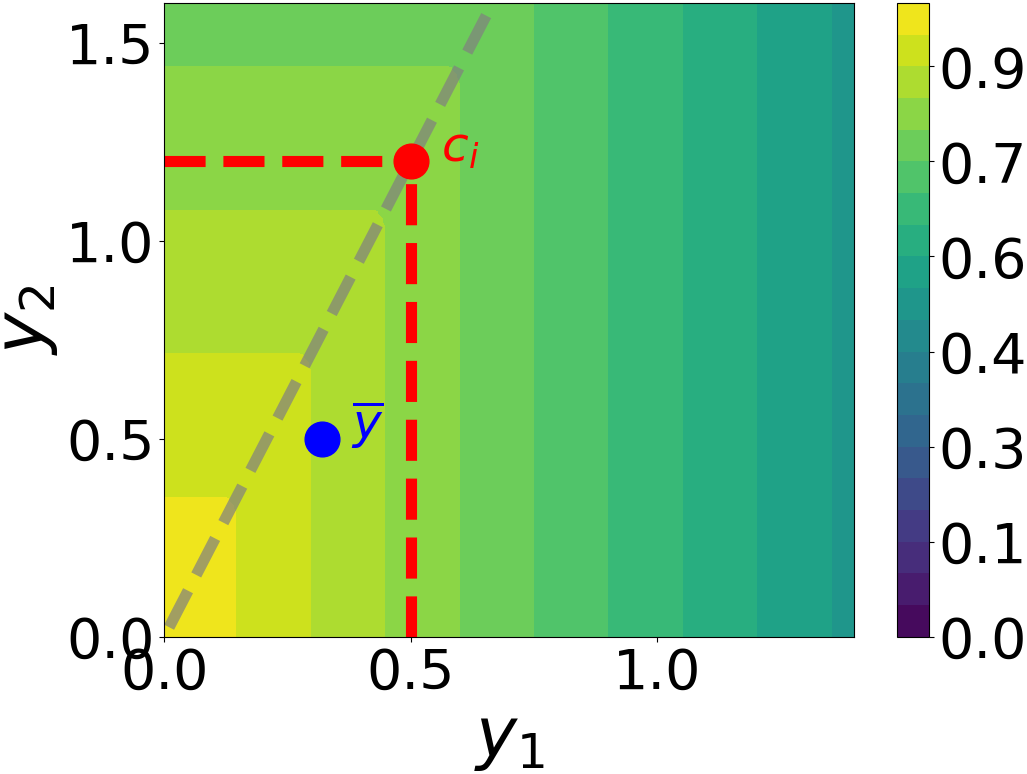}
        \caption{PDUF}
        \label{fig:PDUF}
    \end{subfigure}
    \caption{Contour plots of (a) the commonly used negative $l_1$ distance Utility function (b) the proposed PDUF.}
    \label{fig:combined_figures}
\end{figure}

For instance, the commonly used $\ell_1$ distance (a) fails to satisfy the \textit{Preference Integration} property when calculated from the utopia point, and violates \textit{Dominance Preservation} when calculated from any other point. This is illustrated in Fig.~\ref{fig:l1_utility}, where the contours of an $\ell_1$ distance utility function is shown with an example Pareto-front. Here, the two red points are indistinguishable according to the utility function, demonstrating the limitations of $\ell_1$ distance in distinguishing between Pareto-optimal solutions.

Therefore, we propose the preference-dominated utility function (PDUF) which merges the concept of dominance with user preferences. An illustration of the contours in a 2D case is shown in Fig.~\ref{fig:PDUF}. The PDUF integrates the concept of dominance with user preferences by combining multiple logistic functions centered around different points in the objective function space. 
The PDUF is defined as:
\begin{equation}
\label{eq: PDUF_1}
u(\bm{y}) = \frac{1}{n_c} \sum_{i=1}^{n_c} \prod_{j=1}^{n_y} L_{\beta}(y_j, c_{i,j}),
\end{equation} 

\begin{equation}
\label{eq: PDUF_2}
L_{\beta}(y_j, c_{i,j}) = \frac{1}{1 + \exp\left(\beta \cdot (y_j - c_{i,j})\right)},
\end{equation}
where $c_i = (c_{i,1}, c_{i,2}, \ldots, c_{i,n_y})$ denotes the $i^{\text{th}}$ center for one logistic function,  $\beta$ denotes a parameter that controls the steepness of the logistic function, and $n_c$ denotes the number of centers. 
The logistic function $L_{\beta}(y_j, c_{i,j})$ approximates the step function and enforces dominance for each objective $y_j$, as seen in the red dashed lines in  Fig.~\ref{fig:PDUF}, and the product aggregates this approximation for all objectives. Furthermore, the sum of logistic function products preserve dominance in the objective space. Indeed, for every $\bar{\bm y}$ that dominates user query $\bm c_i$, PDUF will express user preference with $ u(\bar{\bm y}) > u(\bm c_i)$. Finally, the centers define the parameters $\theta_u$ that ensure the utility function adheres to the \textit{preference integration} property by aligning them along an arbitrary line (the grey line in Fig.~\ref{fig:PDUF}). 

\section{Experiments}
\label{sec: experiments}
\begin{figure*}[t]
    \centering
    \includegraphics[width=\textwidth]{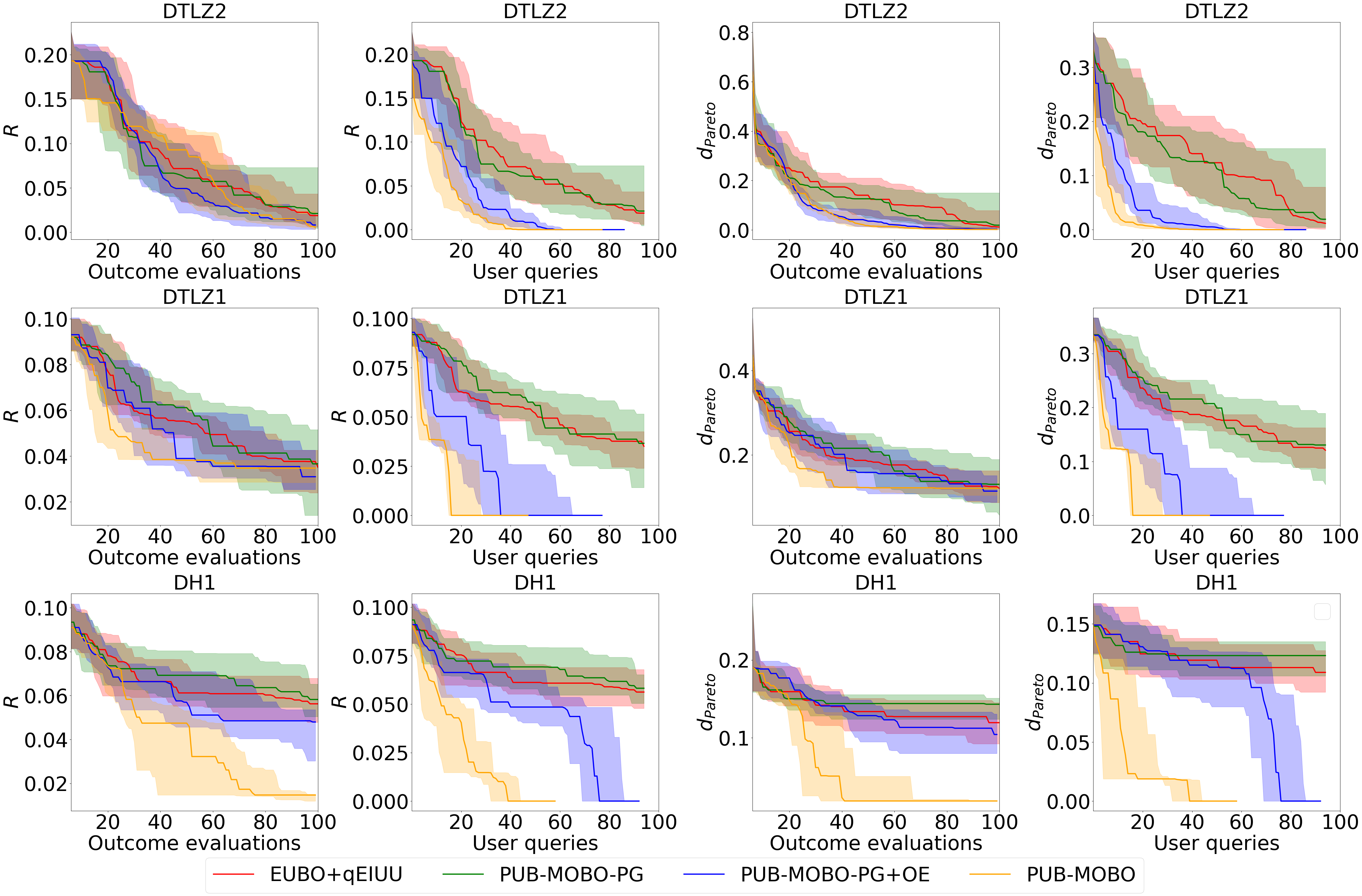}
    \caption{Median and 25-75 percentiles performance comparison on synthetic benchmarks on 20 random seeds with a 100 outcome evaluation budget. Plot titles indicate MOO benchmark; horizontal axis is either number of outcome evaluations or number of user queries. Vertical axis indicates simple regret of utility (user satisfaction) and distance from the Pareto-front (optimality).}
    \label{fig:synthetic}
\end{figure*}
\begin{figure*}[t]
    \centering
    \includegraphics[width=\textwidth]{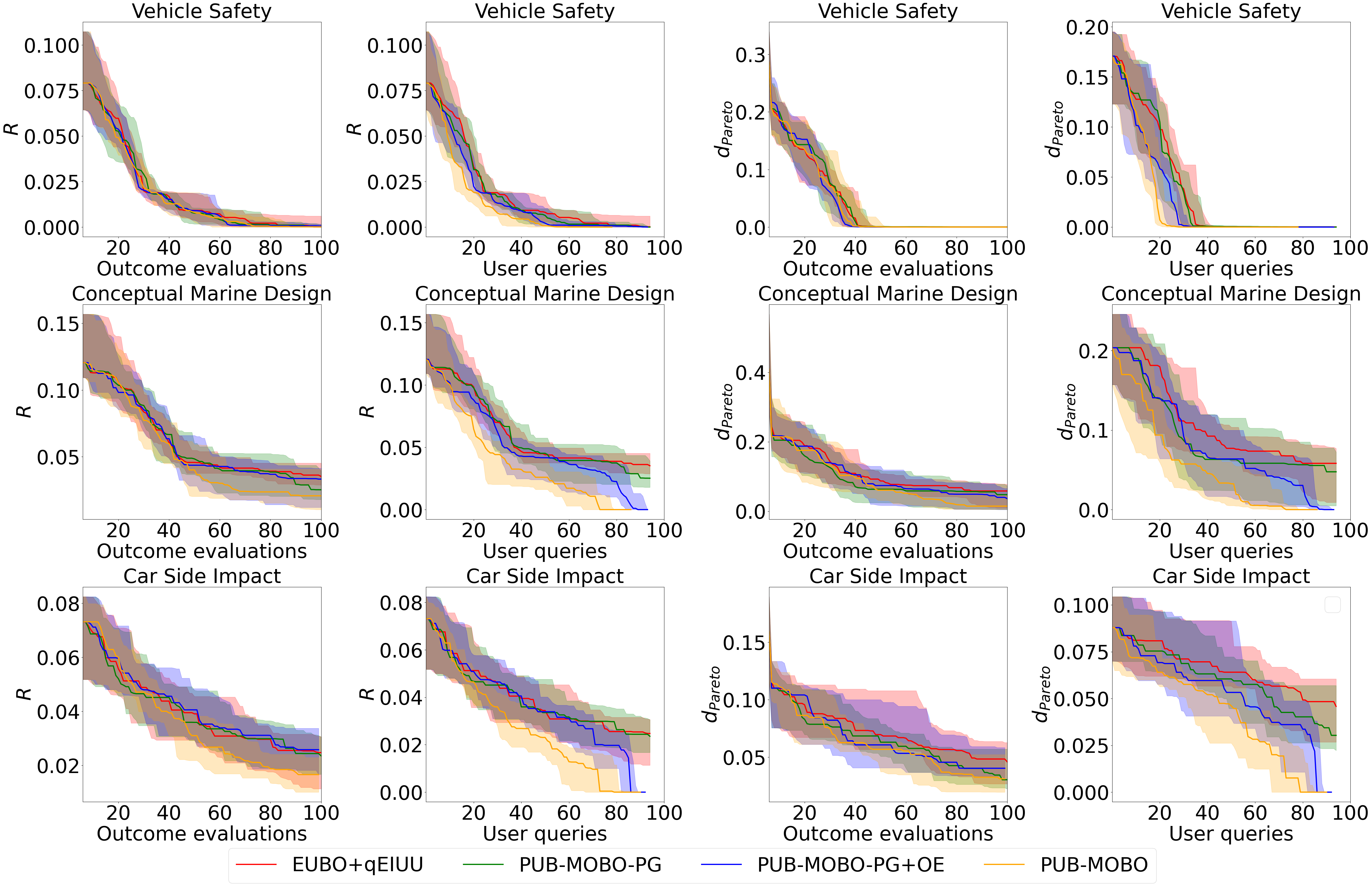}
    \caption{Median and 25-75 percentiles performance comparison on real-world benchmarks. Analogous description  to Fig.~\ref{fig:synthetic}.}
    \label{fig:real}
\end{figure*}
We empirically validate the proposed PUB-MOBO algorithm on 6 benchmark problems and report the performance of utility regret, $R$, and distance to Pareto-front, $d_{\text{Pareto}}$, against outcome evaluations and user queries
\begin{equation}
\label{eq: utility_regret}
R = \frac{u(\bm{f(x^*)}) - u(\bm{f(x)})}{u(\bm{f(x^*)})},
\end{equation}
\begin{equation}
\label{eq: dist_pareto}
d_{\text{Pareto}} 
= \underset{\bm{x}_{\text{Pareto}}\in\mathbb{X}_{\text{Pareto}}}{\text{min}} \left\|\bm{f(x)} - \bm{f(x_{\text{Pareto}})}\right\|_2^2.
\end{equation}

The problems are chosen from synthetic and real-world problems where the Pareto-front is known, so we can evaluate PUB-MOBO in the performance metrics $R$ and $d_{\text{Pareto}}$. All experimental results are obtained using a 13th Gen IntelCore i7-13620H repeated across 20 seeds with hyperparameters $n_{GD} = 10, n_{GI} = 1, \varepsilon = 0.1$. In the experiments, we investigate two aspects of PUB-MOBO: (a) the relevance of using gradient information and (b) the trade-off between cost of outcome evaluations and gradient uncertainty reduction in the GD stage. We compare against the s.o.t.a. preference-based MOBO method \cite{lin2022preference} and two variations of PUB-MOBO:

\begin{itemize}
    \item \eubo \ is a  
    2-stage algorithm proposed in \cite{lin2022preference} which only contains the PE and EXP stages. The acquisition functions used are EUBO and qEIUU in each stage, exactly like in PUB-MOBO. This serves as a baseline of the performance when no gradients are used.
    \item \pubPG\ is a PUB-MOBO ablation that relies solely on predicted gradients (PG) in the GD stage, omitting both outcome evaluations and GI optimizations. This makes it relatively inexpensive, but it ignores the fact that additional samples can yield useful derivative information. This ascertains whether the gradient uncertainty needs to be accounted for at all, and how important the GI step is.
      
    \item \pubPGoe \ is a PUB-MOBO ablation that uses the predicted gradients as in \pubPG, but an Outcome Evaluation (OE) is performed at every gradient descent step in an effort to lower gradient uncertainty around observed points. This further checks whether the expensive GI optimization is needed to improve convergence, or if outcome evaluations are sufficient. 
    
    \item PUB-MOBO is the proposed method from Alg.\ref{alg:PUB-MOBO}-\ref{alg: Local} with up to $n_{\text{GD}}(n_{\text{GI}}+1)$ outcome evaluations in the GD stage. 
\end{itemize}

\subsection{Synthetic Benchmarks}
We examine 3 synthetic problems that are commonly found in MOO literature: DTLZ1, DTLZ2 \cite{deb2005scalable}, and DH1 \cite{deb2005searching}.
The results are displayed in Fig.~\ref{fig:synthetic} in order of increasing $n_x$: DTLZ2 ($n_x$ = 8, $n_f$ = 2), DTLZ1 ($n_x$ = 9, $n_f$ = 2), DH1 ($n_x$ = 10, $n_f$ = 2).

\eubo~is the poorest-performing algorithm across all the synthetic benchmarks, affirming the effectiveness of the additional stage based on local gradient search. The strategy of exploring locally dominating solutions off of the solution proposed by the utility maximization stages, not only results in solutions that are closer to the Pareto-front but also achieves lower utility regret. 
However, PUB-MOBO-PG performs equally poor in most of the synthetic benchmarks, largely due to inaccurate gradient estimation obtained with the surrogate model $ \hat{\bm f}$ by \eqref{eq: dGP_inference_covar}. We frequently observe that the evolution of $\xgd$ in the GD stage is prematurely terminated either due to constraint violations in $\boldsymbol{x}$ or because the Frank-Wolfe algorithm results in $\left\|\bm \alpha^\top\boldsymbol{\mu}^\nabla(\bm x_{\text{GD}})\right\|_2^2\leq\varepsilon_{\text{GD}}$. This occurs not because the algorithm has reached a Pareto-optimal point, but rather due to erroneous gradient estimates. These findings underscore the critical importance of accurate gradient estimation and minimizing gradient uncertainty for the success of PUB-MOBO.
The PG+OE variant significantly outperforms the PG variant in most experiments. Although evaluating the true outcome function at each $\xgd$ incurs additional computational cost, the increased accuracy in gradient estimation from updating the outcome model more than justifies the expense. This leads to a more efficient algorithm overall, as reflected by lower utility regret and closer proximity to the Pareto-front, both in terms of outcome evaluations and user queries.
In a similar trend, PUB-MOBO further improves upon PG+OE variant. The reduction of the gradient uncertainty obtained by evaluating the $\bm x_{\text{GD}}$ suggested by the GI acquisition function, further improve the gradient estimate yielding lower $R$ and $d_{\text{Pareto}}$. This empirically demonstrates that the additional outcome evaluations in the GD stage are justified.

\subsection{Real-World Benchmarks}
We examine 3 problems based on real MOO problems: Vehicle Safety \cite{liao2008multiobjective}, Conceptual Marine Design \cite{parsons2004formulation}, and Car Side Impact \cite{jain2013evolutionary}.
The implementations of these problems are taken from \cite{tanabe2020easy}. The results are shown in Fig.~\ref{fig:real} in order of increasing $n_{x}$: Vehicle Safety ($n_x$ = 5, $n_f$ = 3), Conceptual Marine Design ($n_x$ = 6, $n_f$ = 4), Car Side Impact ($n_x$ = 7, $n_f$ = 4).

Similar to the results on the synthetic benchmarks, the EUBO and PUB-MOBO-PG methods are the worst-performing. The real-world benchmarks present a more complex optimization landscape, which increases the difficulty of accurate gradient estimation. Notably, PUB-MOBO-PG+OE, which performed better on synthetic problems, exhibits similarly poor performance to EUBO and PUB-MOBO-PG on these more challenging benchmarks. In contrast, PUB-MOBO continues to outperform all other methods, consistently delivering near-optimal solutions with respect to both utility regret and proximity to the Pareto-front, due to its effective gradient uncertainty minimization.

The results from both synthetic and real-world experiments demonstrate that incorporating a gradient descent stage in utility-based MOBO leads to solutions with lower utility regret and closer proximity to the Pareto-front, all while requiring fewer outcome evaluations and user queries. Empirical evidence suggests that it is more effective to conduct additional outcome evaluations during each GD stage to achieve accurate gradient steps, as seen in PUB-MOBO, rather than opting for cheaper but less accurate steps. 

\section{Conclusion}
We introduce PUB-MOBO, a sample-efficient multi-objective Bayesian optimization method that integrates user preferences with gradient-based search to find near-Pareto-optimal solutions. Across synthetic and real problems, it achieves high utility and reduced distance to Pareto-front solutions, highlighting the importance of reducing gradient uncertainty in the gradient-based search. Moreover, we introduce a new utility function, PDUF, that respects dominance and models different user preferences. An avenue of future research is to apply the Local Gradient Descent method to other MOBO algorithms.

\bibliography{aaai25}
\end{document}